\pdfoutput=1

\documentclass[11pt]{article}

\usepackage[final]{acl}

\usepackage{times}
\usepackage{latexsym}
\usepackage{microtype}
\usepackage{hyperref}
\usepackage{url}
\usepackage{hhline} 
\usepackage{booktabs}
\usepackage{subcaption}
\definecolor{darkblue}{rgb}{0, 0, 0.5}
\hypersetup{colorlinks=true, citecolor=darkblue, linkcolor=darkblue, urlcolor=darkblue}

\usepackage{algorithm}
\usepackage{makecell}
\usepackage{color}
\usepackage{amssymb}
\usepackage{amsmath}
\usepackage{bbding}
\usepackage{algpseudocode}
\usepackage{algorithmicx,algorithm}

\usepackage{multirow}
\usepackage{booktabs}
\usepackage{graphicx}

\usepackage[T1]{fontenc}

\usepackage[utf8]{inputenc}

\usepackage{microtype}

\usepackage{inconsolata}

\usepackage{graphicx}

\title{Chain-of-Procedure: Hierarchical Visual-Language Reasoning for Procedural QA}

\author{
 \textbf{Guanhua Chen$^1$\thanks{Equal contribution}},
 \textbf{Yutong Yao$^1$\footnotemark[1]},
 \textbf{Shenghe Sun$^1$},
 \textbf{Ci-Jun Gao$^3$},
 \textbf{Shudong Liu$^1$}, \\
 \textbf{Lidia S. Chao$^1$}, 
 \textbf{Feng Wan$^{2,3}$},
 \textbf{Derek F. Wong$^1$\thanks{Corresponding authors.}}
\\
 $^1$NLP\textsuperscript{2}CT Lab, Department of Computer and Information Science,  University of Macau
\\
$^2$Department of Electrical and Computer Engineering, University of Macau 
\\
$^3$Centre for Cognitive and Brain Sciences, University of Macau
\\
\{nlp2ct.guanhua, nlp2ct.yutong, nlp2ct.shenghe, cijun.gao, nlp2ct.shudong\}@gmail.com\\
\{derekfw, lidiasc, fwan\}@um.edu.mo\\
}

\begin{document}
\maketitle
\begin{abstract}
Recent advances in vision-language models (VLMs) have achieved impressive results on standard image-text tasks, yet their potential for visual procedure question answering (VP-QA) remains largely unexplored. VP-QA presents unique challenges where users query next-step actions by uploading images for intermediate states of complex procedures. To systematically evaluate VLMs on this practical task, we propose ProcedureVQA, a novel multimodal benchmark specifically designed for visual procedural reasoning. Through comprehensive analysis, we identify two critical limitations in current VLMs: inadequate cross-modal retrieval of structured procedures given visual states, and misalignment between image sequence granularity and textual step decomposition. To address these issues, we present Chain-of-Procedure (CoP), a hierarchical reasoning framework that first retrieves relevant instructions using visual cues, then performs step refinement through semantic decomposition, and finally generates the next step. Experiments across six VLMs demonstrate CoP's effectiveness, achieving up to 13\% absolute improvement over standard baselines.
\end{abstract}

\section{Introduction}
\begin{figure}
    \centering
    \includegraphics[height=0.5\linewidth]{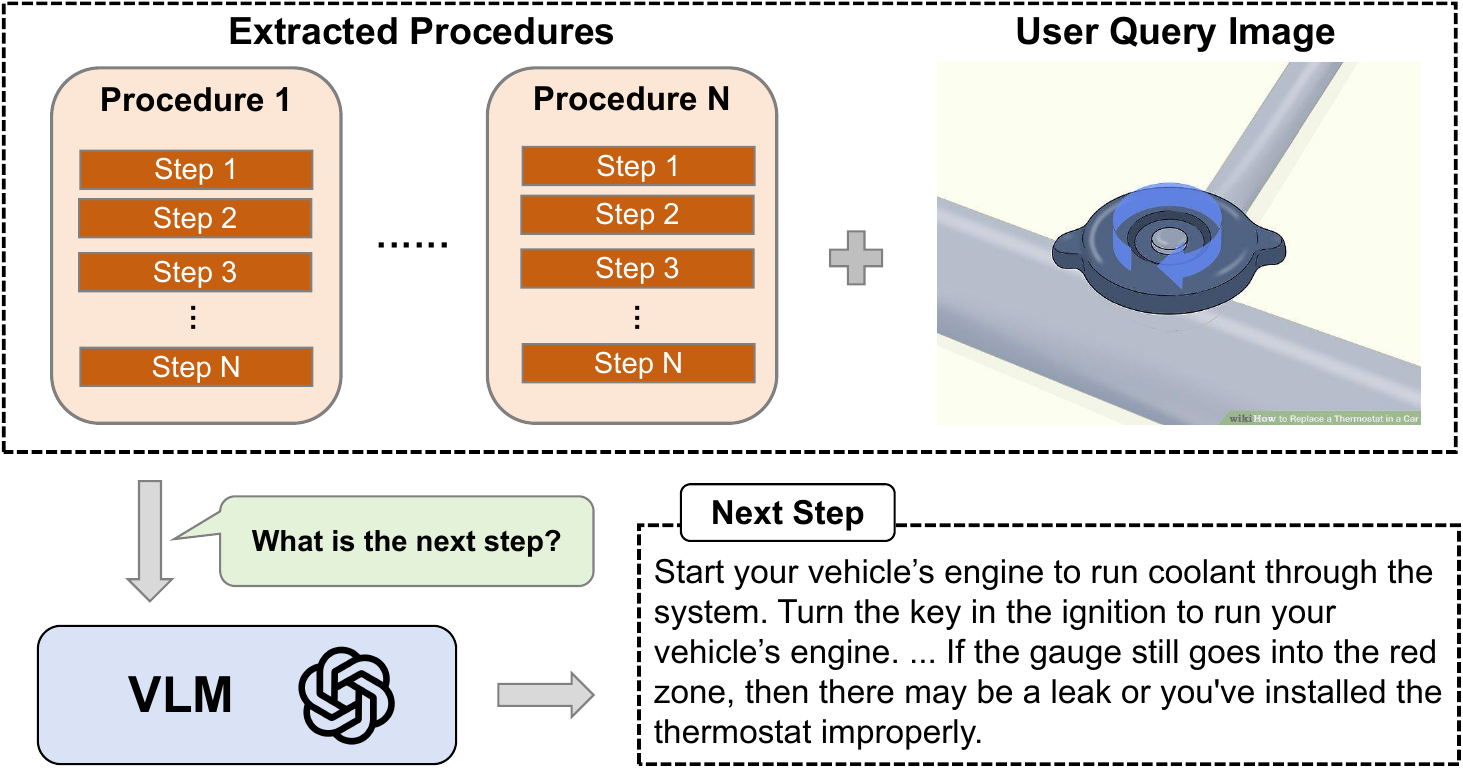}
    \caption{An example of the next-step prediction task.}
    \label{fig:7}
\end{figure}

Recent Large Language Models (LLMs) such as GPT-4 \cite{Hurst2024GPT4oSC} and Claude \cite{claude-3.7} have demonstrated strong performance across various multimodal understanding tasks \cite{ma2025visaidmath}, including visual question answering (VQA) \cite{qiu2024snapntell, nguyen2025owlviz} and retrieval-augmented generation (RAG) \cite{chen-etal-2025-sgic,DBLP:conf/iclr/YuTXCRYLWHL025}. However, their capability in visual procedure question answering (VP-QA), a critical competency for applications like furniture assembly, device troubleshooting, and cooking assistance, remains largely underexplored. We formally define this task as a multimodal selection problem: given an image representing an intermediate state of a procedure for a complete task workflow, the model must reason about the progress and identify the correct next step (a major action in the procedure) from a set of candidate steps. As illustrated in Figure \ref{fig:7}, when a user uploads an image of a partially disassembled radiator cap, the system must align this visual state with the specific step of the procedure to predict the immediate next action. This task requires fine-grained visual-textual alignment and hierarchical reasoning, presenting unique challenges that surpass those of conventional image-text tasks.

While current Vision-Language Models (VLMs) focus on static visual QA \cite{bazi2023vision,guo2023images} and text-guided image reasoning \cite{kelly2024visiongpt, DBLP:conf/eccv/ZhouHWWW24}, procedural understanding requires a dynamic alignment between visual queries and structured plans. Existing benchmarks either focus on single-image recognition \cite{yang2021visual} or rely on pre-aligned procedural data \cite{lin2020recipe}, neglecting the core challenge of grounding evolving visual states to hierarchical procedure. Unlike text-guided document reasoning \cite{yang2018hotpotqa}, it is significantly more challenging for VLMs to align visual states with discrete textual instructions, particularly when the granularity of visual evidence does not perfectly match the text.

To fill these gaps, we construct \textbf{ProcedureVQA}, a novel benchmark for visual procedural step understanding. We reconstruct a released dataset proposed by \citet{chen2024comm} into a multimodal benchmark spanning  five common procedural domains (home repair, cooking, device setup, crafts, and health care). We evaluate six state-of-the-art VLMs on end-to-end 
next-step prediction and observe suboptimal accuracy (25.1-62.5\%). Through error analysis, we identify a critical "granularity mismatch": high-level instructions often encompass multiple implicit atomic actions, making direct visual alignment difficult for current models.

Consequently, we propose \textbf{Chain-of-Procedure (CoP)}, a hierarchical reasoning framework designed to bridge this gap. Unlike prior retrieval-augmented approaches \cite{panda-etal-2024-holmes} that treat procedures as flat text, CoP operates in three distinct phases to align terminology with visual evidence. First, it employs VLMs to filter broadly relevant candidate procedures. Second, it dynamically decomposes the coarse procedure into fine-grained sub-steps to match the visual state granularity. Finally, it predicts the current sub-step action and infers the precise next step. By decoupling procedure retrieval, step decomposition, and visual-textual alignment, CoP significantly strengthens VLMs' procedural reasoning capabilities. We evaluate CoP on six VLMs using our ProcedureVQA benchmark, and experimental results demonstrate that CoP improves the performance of next-step prediction by 1.7-17.9\% in BertScore and 2.2-13\% in LLM-Score over baselines.

\section{Related Work}

\subsection{Procedural Reasoning and Planning}
Procedural reasoning involves understanding goal-oriented sequences, tracking progress, and anticipating future actions. Early text-based benchmarks like PIQA \cite{bisk2020piqa} focused on physical commonsense reasoning, while ProPara \cite{dalvi2018tracking} and OpenPI \cite{tandon2020dataset} introduced entity state tracking within procedural texts. The field expanded with large-scale instructional video datasets such as COIN \cite{DBLP:conf/cvpr/TangDRZZZL019}, HowTo100M \cite{miech2019howto100m}, and Ego4D \cite{DBLP:conf/cvpr/GraumanWBCFGH0L22}, enabling tasks like weakly supervised procedure learning \cite{Zhukov2019CrossTaskWS}. RecipeQA \cite{yagcioglu-etal-2018-recipeqa} further explored multimodal comprehension over recipe procedures. Recent work has pivoted towards multimodal procedural planning: \citet{DBLP:conf/emnlp/LuL0Z0W24} utilizes dual text-image prompting to generate coherent plans, while \citet{DBLP:conf/emnlp/Gloria-SilvaSM24} grounds instructional plans within visual environments. However, these approaches predominantly rely on continuous video temporal cues, sequential texts, or paired image-text contexts. In contrast, ProcedureVQA targets snapshot-based procedural reasoning, requiring models to infer implicit progress and predict next-step actions from a single static intermediate visual state.

\subsection{Multimodal Learning for Procedure Task}
The application of VLMs to specific procedure domains, particularly cooking and maintenance, has garnered significant attention. \citet{kafle2017visual} and \citet{yagcioglu-etal-2018-recipeqa} laid the groundwork with datasets like RecipeQA. More recent efforts have developed specialized models for these tasks. CookingQA \cite{Khilji2021CookingQAAQ} addresses QA based on ingredients, while FoodLMM \cite{Yin2023FoodLMMAV} and LLaVA-Chef \cite{mohbat2024llava} leverage large multimodal models to generate comprehensive recipes and cooking procedures. Furthermore, retrieval-based approaches have been explored to enhance recipe understanding \cite{Song2023EnhancingRR}.
Despite these advancements, a key limitation persists: the misalignment between the granularity of visual states and textual procedures. While general VLMs \cite{liu2023visual, zhao2024mg} show promise, they struggle with the structured retrieval required for complex procedures. Unlike previous works that focus on end-to-end generation or simple alignment, our CoP framework explicitly tackles the granularity mismatch through semantic step decomposition, enabling more accurate next-step predictions.

\section{ProcedureVQA} \label{sec.3}
We present a novel benchmark reconstructed from the WikiHow portion of the CoMM dataset released by \citet{chen2024comm}. Our framework targets the broad spectrum of real-world procedural tasks, ranging from technical vehicle maintenance to daily lifestyle skills.

\subsection{Problem Formulation} 
To rigorously define the task and standardize terminology, we formalize VP-QA as follows. A procedure $\mathcal{P} = \{S_1, S_2, ..., S_L\}$ is a complete task workflow composed of $L$ sequential steps (major actions). Given a visual query $v_t$ showing the intermediate state after completing step $S_t$, the model must: (1) retrieve the correct procedure $\mathcal{P}^*$ from a candidate set $\mathcal{C}_p = \{\mathcal{P}_1, ..., \mathcal{P}_N\}$ containing one correct procedure and $N-1$ negative procedures, (2) localize $v_t$ to the current step $S_t$ within $\mathcal{P}^*$, and (3) predict the content of the next step $S_{t+1}$. This formulation tests cross-procedure retrieval, visual state grounding, and next-step reasoning under granularity mismatch between coarse-grained steps and fine-grained visual evidence.

\begin{table}[t]
  \centering
    \renewcommand\arraystretch{1}
    \tabcolsep=0.12cm
    \scalebox{1}{
    \begin{tabular}{lccc}
   \toprule
    \textbf{Metric} & \textbf{Train} & \textbf{Test} & \textbf{Total}  \\
    \midrule
    Total Samples & 1,939 & 2,844 & 4,783  \\
    Avg. Length (tokens) & 5.9 & 5.7 & 5.8  \\
    Step Length Std & 7.1 & 4.8 & 5.9  \\
    Shortest Sample & 3 & 3 & 3  \\
    Longest Sample & 45 & 22 & 33.5  \\
    \midrule
    \multicolumn{4}{c}{\textbf{Domain Distribution}} \\
    \midrule
     Cars \& Other Vehicles & 476 & 509 & 985  \\
    Computers \& Electronics & 696 & 631 & 1,327  \\
    Hobbies \& Crafts & 551 & 533 & 1,084 \\
    Sports \& Fitness & 216 & 500 & 716 \\
    Work World & 0 & 671 & 671 \\
    \bottomrule
    \end{tabular}}
    \caption{The statistics of the ProcedureVQA dataset consist of benchmarks and a training set.}
  \label{table1}
\end{table}

\subsection{Data Construction} \label{sec.3.2}
Our reconstruction framework transforms raw procedural data into a multimodal reasoning benchmark through domain-specific processing. 

\paragraph{Data Domains.} We strategically selected five domains to ensure task diversity while maintaining domain coherence: (1) \textit{Cars \& Vehicles} covers maintenance tasks like oil checks; (2) \textit{Computers \& Electronics} includes hardware and software tutorials; (3) \textit{Hobbies \& Crafts} features visual guides on arts; (4) \textit{Sports \& Fitness} provides exercise guides; and (5) \textit{Work World} covers professional skills. This selection balances everyday domains with technically complex ones. The example of each domain is shown in Appendix \ref{app.3}.

\paragraph{Granularity Simulation via Step Fusion.}
A key challenge in procedural VQA is the inconsistency between visual and textual granularity. To simulate this real-world variation, we apply controlled step fusion to the raw textual instructions. With a 50\% probability, two consecutive atomic steps are merged into a single coarse-grained instruction. This approach, inspired by \citet{Gao2022SummarizingPT} and \citet{dalvi2018tracking}, creates a realistic mismatch where a single step may correspond to multiple visual states. This design is crucial for testing a model's robustness to inconsistent step resolutions and motivates the need for step decomposition. We also conduct a sensitive analysis of this probability in Appendix \ref{app.9}.

\paragraph{Negative Sampling and Data Structure.}
To create challenging distractors, we encode all images into 512-dimensional vectors via ResNet50 \cite{he2016deep}. For each input image $v_t$, we curate hard-negative candidates through intra-domain cosine similarity searches. We select visually proximate but procedurally irrelevant instructions as distractors to prevent the model from relying on simple visual shortcuts. 
Each data instance is finally constructed as a tuple:
\begin{equation}
    D = (v_t, \mathcal{C}, y)
\end{equation}
where $v_t$ is the input image, $\mathcal{C} = \{c_{pos}, c_{neg}^{(1)}, c_{neg}^{(2)}\}$ is the set of three candidate instructions (one positive, two hard negatives), and $y$ is the index of the ground truth. This structure ensures that the benchmark rigorously evaluates the VLM's ability to perform precise image-to-text alignment within a procedural context.

\begin{figure}
    \centering
    \includegraphics[height=0.65\linewidth]{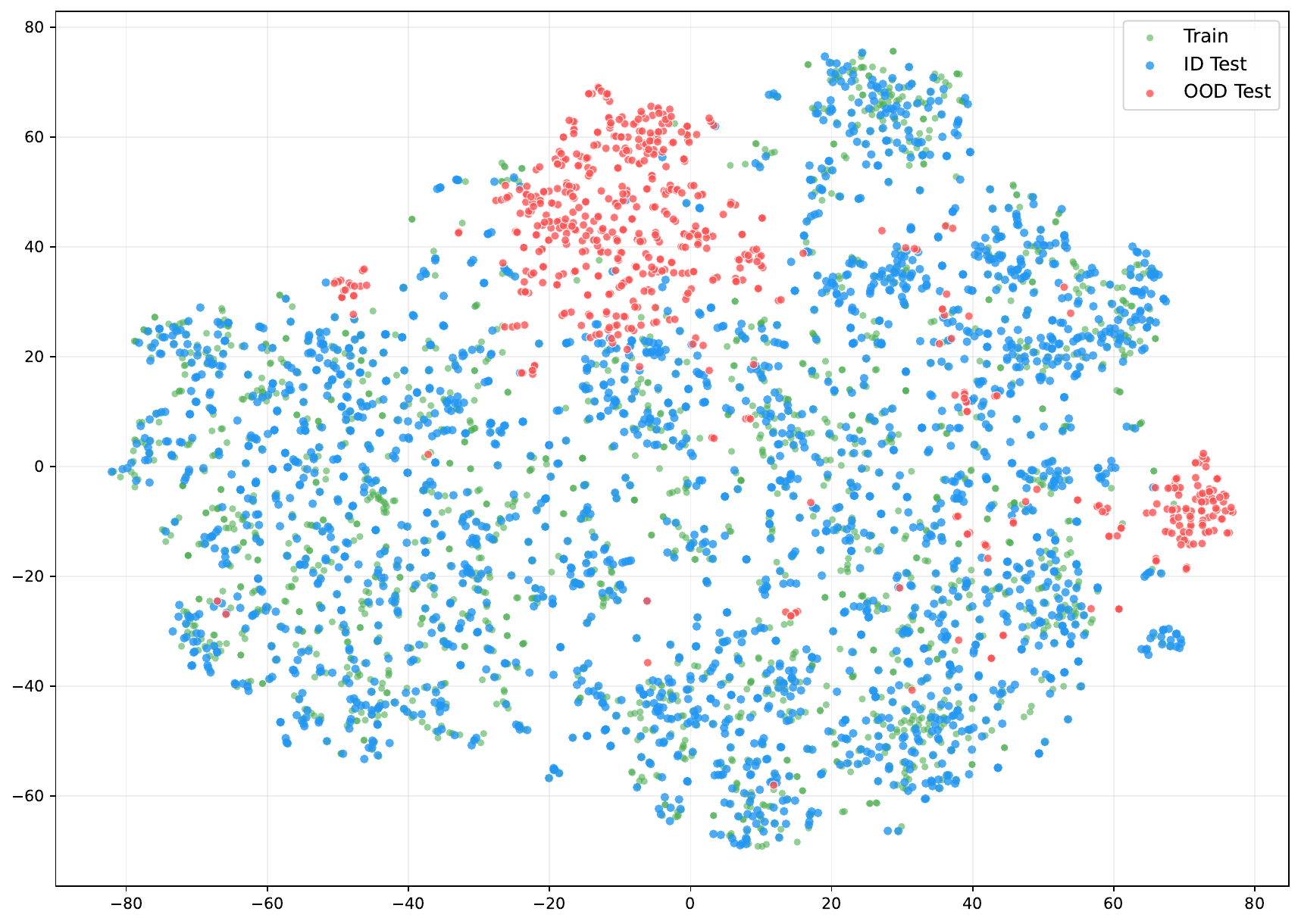}
    \caption{The t-SNE of the training data and in-domain (ID)/out-of-domain (ODD) test data.}
    \label{fig:3}
\end{figure}

\subsection{Data Statistics} \label{sec.3.3}
Our dataset contains 4,783 samples. To evaluate VLM fine-tuning, we split the data into training and benchmark sets: all ``Work World'' 671 samples are reserved as out-of-distribution (OOD) tests, while other domains use step-length-stratified splits for balanced complexity. As Table \ref{table1} shows, training and in-domain (ID) benchmarks share similar lexical complexity, but training exhibits higher step variance, retaining long-tail sequences.

We measure semantic overlap using \textbf{all-roberta-large-v1} \cite{DBLP:conf/emnlp/ReimersG19}. Benchmark-training pairs show a median cosine similarity of 0.64, balancing transfer potential and memorization resistance. The t-SNE visualizations in Figure \ref{fig:3} confirm structural alignment: training (green) and ID test data (blue) overlap significantly, while OOD samples (red) form distinct clusters. This design rigorously tests the learning and generalization capabilities of VLMs on VP-QA task.


\section{Benchmarking and Analysis} \label{sec.4}
This section evaluates the capabilities of powerful VLMs on VP-QA task using the \textbf{ProcedureVQA} benchmark. We first introduce the experimental setup and evaluation metrics, followed by a comprehensive analysis of benchmark results and five decomposed fundamental sub-tasks to enable systematic analysis of VLM behaviors.

\begin{table}[t]
  \centering
    \renewcommand\arraystretch{1}
    \tabcolsep=0.38cm
    \scalebox{1}{
    \begin{tabular}{lcc}
    \toprule
    \textbf{Model} & \multicolumn{1}{c}{\textbf{Zero-Shot}}  & \multicolumn{1}{c}{\textbf{CoT}} \\
      \midrule
      Qwen2.5-VL-7B  & 25.1  & 26.6 \\
      Qwen2.5-VL-72B  & 43.4 & 41.3   \\
      \midrule
      GPT-4o-mini  & 34.4  & 42.7   \\
      GPT-4o  & \textbf{62.5}  & 51.8  \\
      Gemini-2.5-Flash & 44.7  & 42.6 \\
      Claude-3-5-Sonnet & 47.5  & \textbf{55.6}   \\
      
      \bottomrule
    \end{tabular}}
    \caption{The results (\%) on the benchmarks of different VLMs. \textbf{Bold} indicate the best result for each setting.}
  \label{table2}
\end{table}

\subsection{Models and Evaluation Metrics}
We evaluate two categories of VLMs: open-source (Qwen2.5-VL-7B and Qwen2.5-VL-72B) \cite{Qwen-VL} and closed-source VLMs (GPT-4o-mini, GPT-4o, Gemini-2.5-Flash, Claude-3-5-Sonnet) \cite{DBLP:journals/corr/abs-2303-08774, DBLP:journals/corr/abs-2403-05530, DBLP:journals/corr/abs-2407-01557}. The implementation details of all models are explained in Appendix \ref{app.2}. For evaluation, we employ \textbf{Accuracy} as the primary metric to measure whether the predicted next-step action matches the ground truth. This straightforward metric allows us to assess the model's fundamental capability on this task. Two prompting strategies are compared: zero-shot prompting and chain-of-thought (CoT) prompting \cite{DBLP:conf/nips/Wei0SBIXCLZ22}.

\begin{figure}
    \centering
    \includegraphics[width=1\linewidth]{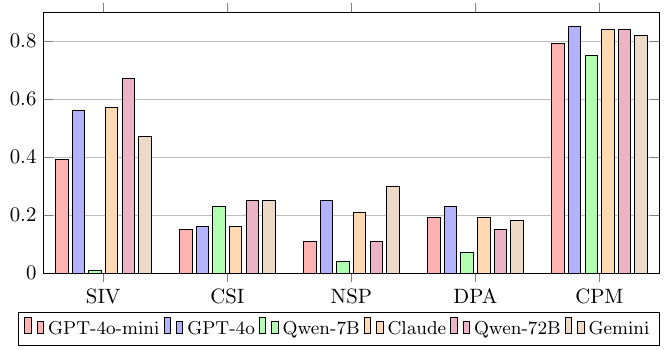}
    \caption{The experimental results (Accuracy) of our designed five different sub-tasks on six powerful VLMs.}
    \label{fig:1}
\end{figure}

\subsection{Performance Across Different VLMs}
Table \ref{table2} presents the benchmarking results across zero-shot and CoT prompting. Closed-source VLMs consistently outperform open-source models, with GPT-4o achieving the highest zero-shot accuracy of 62.5\%. CoT prompting shows mixed effects on accuracy: while it improves performance for some models, Qwen2.5-VL-72B experiences a 2.1\% accuracy decline, potentially due to over-elaboration in step generation. Model scale plays a critical role in performance, as the 72B variant of Qwen2.5-VL demonstrates an 18.3\% accuracy advantage over its 7B counterpart. Despite these advancements, even the most powerful VLMs struggle to exceed 65\% accuracy, highlighting persistent challenges in image-guided procedural reasoning.

\begin{figure*}
    \centering
    \includegraphics[width=1\linewidth]{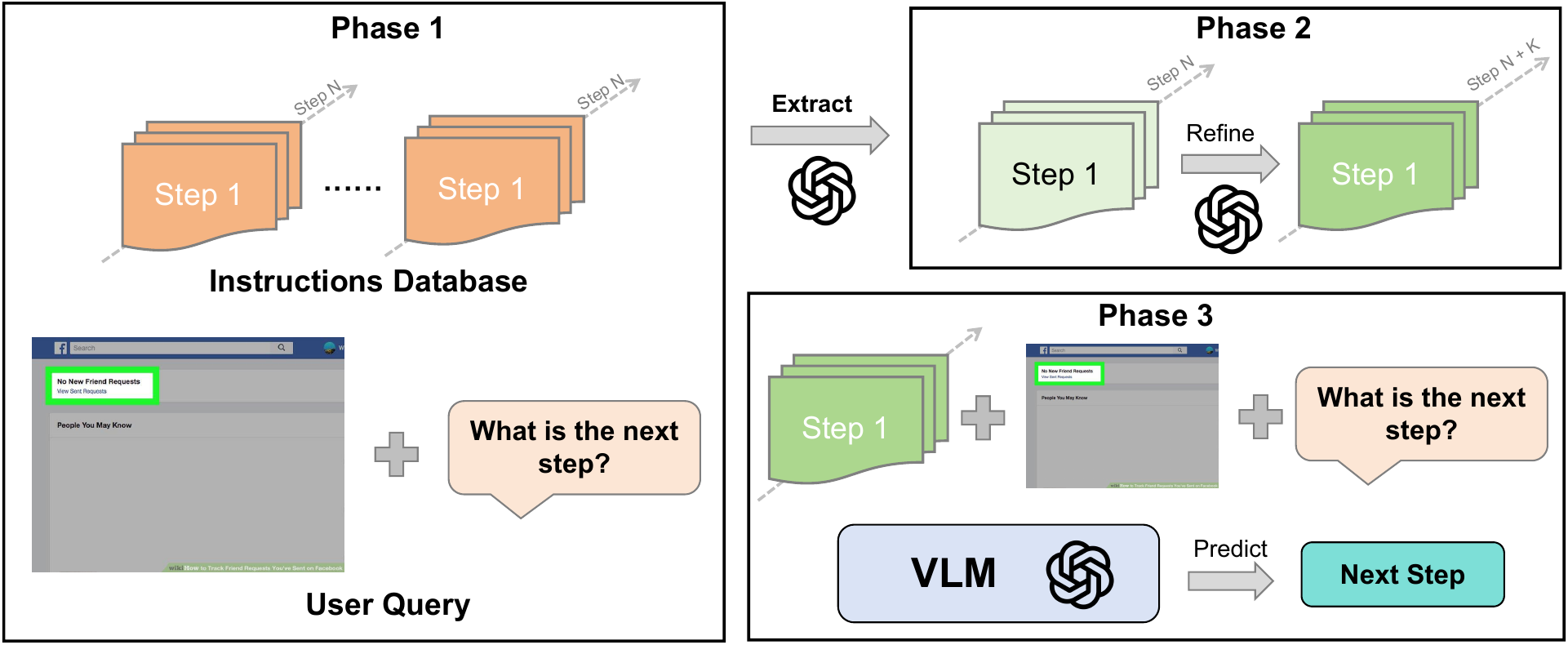}
    \caption{The overview of our proposed framework.}
    \label{fig:4}
\end{figure*}

\subsection{Decomposing Task-Specific Abilities} \label{sec.4.3}
To systematically investigate procedural reasoning capabilities in VLMs, we design five diagnostic sub-tasks with increasing complexity (detailed task formulations in Appendix \ref{appendix:task_details}). Results are shown in Figure \ref{fig:1}. We begin with Shuffled Procedure Verification (SIV), where models determine if procedure steps are correctly ordered—Qwen2.5-VL-72B achieves 67.2\%, surpassing Claude-3.5-Sonnet. Current Step Identification (CSI) requires recognizing which step corresponds to a given visual state, testing temporal grounding. Next-Step Prediction (NSP) extends this by predicting the subsequent action, showing a 3\% performance drop from CSI, indicating increased difficulty in forward temporal reasoning. Disordered Procedure Adaptation (DPA) simulates real-world scenarios by requiring step continuation from shuffled procedures. Finally, Cross-Procedure Matching (CPM) evaluates whether an image-procedure pair are semantically aligned, where GPT-4o achieves state-of-the-art performance (84.5\%).

Three critical findings emerge. First, VLMs excel at structural validation tasks (SIV and CPM: 78.9\% and 75.6\% average accuracy) but struggle with procedural reasoning tasks (NSP and DPA), revealing that multimodal alignment capabilities exceed causal reasoning proficiency. Second, the 23.9\% performance gap between CSI and NSP highlights VLMs' fundamental difficulty in reasoning about future actions versus recognizing current states. Third, closed-source models show 15.2\% smaller accuracy degradation on DPA compared to open-source alternatives, demonstrating superior robustness to procedural perturbations. These findings motivate CoP's design to explicitly address the challenge of VP-QA task.

\section{Method}
Building upon the insights from Section \ref{sec.4.3}, we present Chain-of-Procedure (CoP), a framework that redefines VP-QA through structured hierarchical reasoning. Figure \ref{fig:4} depicts CoP coordinates VLMs across three tightly coupled phases: context-aware procedure retrieval, step-wise procedural decomposition, and next-step prediction. This framework tackles key limitations of monolithic VLMs by explicitly decoupling three core aspects: cross-modal alignment, fine-grained procedural reasoning, and temporal state modeling.

\subsection{Context-Aware Procedure Retrieval}
The initial phase resolves the critical challenge of extracting the procedure relevant to the input image. Instead of relying on static embeddings, we operationalize this retrieval as an iterative cross-modal matching process. Motivated by Section \ref{sec.4.3}, we design a verification loop where the VLM iteratively evaluates candidate procedures against the input image. Specifically, consider a database containing N procedures: [Procedure 1], [Procedure 2], ..., [Procedure N]. For each procedure, the VLM computes a multimodal alignment score using the format "Task Instruction + [Procedure] + Image Query", where the instruction prompts the VLM to provide a score addressing the core question: "Does this procedure accurately reflect the image's content?" The highest-scoring procedure is dynamically selected and used in Phase 2 of our method, replacing static retrieval with an adaptive grounding process that leverages VLMs' intrinsic cross-modal reasoning while ensuring computational efficiency.

\subsection{Step-Wise Procedural Decomposition}
A core innovation of CoP lies in its explicit modeling of procedural granularity, a dimension severely underexploited by conventional VLMs. We design a decomposition module that reconstructs atomic procedural steps from potentially over-grouped actions. The VLM first identifies composite steps through syntactic patterns and semantic meaning, then splits them into temporally ordered atomic actions while keeping unnecessary steps unchanged. Importantly, this process preserves the original procedure’s causal structure while recovering intermediate states omitted in aggregated steps. 

\subsection{Next Step Prediction}
In the final phase, our method establishes cross-modal alignments between the input image and the corresponding procedural step. As demonstrated in Section \ref{sec.4.3}, VLMs exhibit superior capability in identifying the current state of an input. Thus, our approach first reasons about the input’s present state before inferring subsequent steps, rather than directly predicting the next step. Specifically, the VLM implicitly computes a multimodal alignment score between the image and each decomposed step. The model then outputs the identified current state, and implicitly predicts the next step.

Through these three phases, we establish a new paradigm for hierarchical procedural reasoning for VLMs. All the detailed instruction templates for VLMs are explained in Appendix \ref{app.5}.

\begin{table}[t]
  \centering
    \renewcommand\arraystretch{1}
    \tabcolsep=0.16cm
    \scalebox{1}{
    \begin{tabular}{lcc}
    \toprule
     Model &  BertScore &  LLM-Score   \\
     \midrule
     \multicolumn{3}{c}{\textbf{ProcedureVQA}} \\
      \midrule
      Qwen2.5-72B  &  30.1 & 40.6  \\
       Qwen2.5-72B (Ours) & \textbf{51.7} & \textbf{49.6} \\
      \midrule
      GPT-4o-mini  &  38.6 & 40.2 \\
       GPT-4o-mini (Ours) & \textbf{48.3} & \textbf{48.7} \\
       \midrule
      GPT-4o  &  43.5 & 54.2 \\
      GPT-4o (Ours) &  \textbf{61.4} &\textbf{67.2} \\
      \midrule
      Gemini-2.5 &  46.4 &  43.6 \\
      Gemini-2.5 (Ours) &  \textbf{48.1} & \textbf{47.3} \\
      \midrule
      Claude-3-5 & 44.7 & 45.6 \\
      Claude-3-5 (Ours) & \textbf{48.8} & \textbf{47.8} \\
      \midrule
      \multicolumn{3}{c}{\textbf{RecipeQA}} \\
      \midrule
      Qwen2.5-72B  &  23.9 & \textbf{35.0}  \\
       Qwen2.5-72B (Ours) & \textbf{26.5} & 32.3 \\
      \midrule
      GPT-4o-mini  &  20.7 & \textbf{31.9} \\
       GPT-4o-mini (Ours) & \textbf{22.0} & 29.6 \\
       \midrule
      GPT-4o  &  38.7 & 48.6 \\
      GPT-4o (Ours) &  \textbf{46.6} & \textbf{49.7} \\
      \midrule
      Gemini-2.5 &  \textbf{42.8} & \textbf{47.2} \\
      Gemini-2.5 (Ours) &  37.8 & 44.6 \\
      \midrule
      Claude-3-5 & 16.8 & 26.3  \\
      Claude-3-5 (Ours) & \textbf{19.3} & \textbf{33.6} \\
      
      \bottomrule
    \end{tabular}}
    \caption{The main results (\%) of our CoP framework on the ProcedureVQA and RecipeQA datasets. \textbf{Bold} numbers indicate the better result for each baseline.}
  \label{table3}
\end{table}

\begin{table}[t]
  \centering
    \renewcommand\arraystretch{1}
    \tabcolsep=0.28cm
    \scalebox{1}{
    \begin{tabular}{lcc}
    \toprule
     Model& BertScore & LLM-Score   \\
      \midrule
      Baseline & 59.7& 67.4  \\
      End2End (Ours) &  61.2& 69.2 \\
      Pipeline (Ours) & \textbf{63.6} & \textbf{72.1} \\  
      \bottomrule
            \end{tabular}}
    \caption{The results of fine-tuning Qwen2.5-VL-7B on ProcedureVQA dataset with three different setting. \textbf{Bold} numbers indicate the best result. }
  \label{table11}
\end{table}

\section{Experiments and Analysis} \label{sec.5}
\subsection{Setup}
To validate the effectiveness of the proposed CoP framework, we conducted comprehensive experiments across six VLMs following Section \ref{sec.4}. Moreover, we evaluate the performance of the fine-tuned Qwen2.5-VL-7B model. Besides our proposed ProcedureVQA dataset, we also evaluate our CoP on the RecipeQA dataset \cite{yagcioglu-etal-2018-recipeqa} to verify the generalizability of our CoP. For deep analysis, we employ two different evaluation metrics: BERTScore \cite{zhang2019bertscore} and LLM-Score, which are detailed in Appendix \ref{app.1}.

\subsection{Main Results}

Table \ref{table3} show the main results of our CoP framework on two benchmarks. On the ProcedureVQA dataset, our method achieves consistent improvements across all models. Notably, Qwen2.5-72B with CoP obtains a 21.6\% absolute improvement in BERTScore and 9.0\% in LLM-Score. GPT-4o also shows substantial gains, reaching 61.4\% BERTScore and 67.2\% LLM-Score. These results demonstrate that our framework effectively enhances the alignment between visual understanding and procedural reasoning.

On the RecipeQA dataset, the improvements are more varied. GPT-4o and Claude-3-5 benefit clearly from our method, with GPT-4o achieving 7.9\% improvement in BERTScore and Claude-3-5 gaining 7.3\% in LLM-Score. However, Gemini-2.5 shows decreased performance on this dataset. We attribute this to the inherent differences between the two benchmarks. RecipeQA contains more diverse and less structured cooking instructions compared to ProcedureVQA. Since Gemini-2.5 already achieves strong baseline performance on RecipeQA, the additional constraints from CoP may limit its flexibility in generating varied responses. Similarly, the slight LLM-Score decreases for Qwen2.5-72B and GPT-4o-mini suggest that these LLMs may over-adapt to the structured output format, which is less suitable for the open-ended nature of recipe descriptions.

We also compare different fine-tuning strategies in Table \ref{table11}. Specifically, we examine the performance of our step-wise LoRA fine-tuning against fine-tuning with CoT~\cite{wei2022chain} style synthetic data that follows our pipeline. The results show that both fine-tuning methods outperform the direct fine-tuning baseline. While the CoT fine-tuning approach offers faster inference speed, it achieves lower performance compared to our step-wise fine-tuning method. Details of training data construction are provided in Appendix \ref{app:7}.

\begin{table}[t]
  \centering
    \renewcommand\arraystretch{1}
    \tabcolsep=0.16cm
    \scalebox{1}{
    \begin{tabular}{lcc}
    \toprule
     Model& BertScore & LLM-Score   \\
      \midrule
      \multicolumn{3}{c}{ProcedureVQA} \\
      \midrule
      GPT-4o  & 43.5 & 54.2 \\
      GPT-4o (Phase 1)  & 47.3 & 62.6  \\
      GPT-4o (Phase 1\&2)  &   47.1  &  52.6 \\
      GPT-4o (Phase 1\&3)  &  59.8  & 66.8  \\
      GPT-4o (Ours) & \textbf{61.4}  &  \textbf{67.2}\\    
      \midrule
       \multicolumn{3}{c}{Recipe} \\
      \midrule
      GPT-4o  & 38.7 & 48.6 \\
      GPT-4o (Phase 1)  & 38.8 & 47.7  \\
      GPT-4o (Phase 1\&2)  &  38.5  &  47.1 \\
    GPT-4o (Phase 1\&3)  &  45.5 & 48.7  \\
      GPT-4o (Ours) & \textbf{46.6} & \textbf{49.7} \\      
      \bottomrule
    \end{tabular}}
    \caption{The ablation study of our method on two datasets with GPT-4o. \textbf{Bold} numbers indicate the better result for each baseline.}
  \label{table4}
\end{table}

\subsection{Ablation Studies}
To investigate the necessity of each component in our framework, we conduct an ablation study on the GPT-4o model with two datasets. The experimental results are shown in Table \ref{table4}. Our analysis progressively integrates components from phase 1 through phase 3. The results show that combining all three phases achieves the best performance: 61.4 BertScore and 67.2 LLM-Score on ProcedureVQA, and 46.6 BertScore and 49.7 LLM-Score on RecipeQA. Interestingly, adding phase 2 to phase 1 causes a performance drop (from 62.6 to 52.6 LLM-Score on ProcedureVQA), suggesting that phase 2 may introduce noise during procedure decomposition. However, phase 3 effectively addresses this issue: phase 1\&3 substantially outperforms phase 1 alone, and our complete framework further improves upon Phase 1\&3. This demonstrates that while phase 2 alone may be suboptimal, the combination of all three phases can achieve optimal performance.

\begin{table}[t]
  \centering
    \renewcommand\arraystretch{1}
    \tabcolsep=0.38cm
    \scalebox{1}{
    \begin{tabular}{lcc}
    \toprule
     Model& BertScore & LLM-Score   \\
      \midrule
      Baseline & 43.5 & 54.2 \\
      CLIP-Full & 29.1 & 34.5 \\
      CLIP-P1 & 47.5 & 51.6  \\
      CLIP-P3 & 37.9  & 45.0 \\  
      Ours & \textbf{61.4} & \textbf{67.2} \\
      \bottomrule
            \end{tabular}}
    \caption{The results of replacing LLM with CLIP Embedding model for different phases on GPT-4o. \textbf{Bold} means the best result. }
  \label{table12}
\end{table}

\subsection{Comparison with Retrieval Models}
To further validate the necessity of our approach, we compare against CLIP\footnote{https://huggingface.co/openai/clip-vit-large-patch14}~\cite{radford2021learning}, a strong multimodal retrieval model that maps image and text into a shared embedding space for similarity computation. We evaluate CLIP in three configurations: (1) CLIP-P1: replacing only Phase 1 with CLIP-based retrieval while keeping LLM-based generation; (2) CLIP-P3: replacing only Phase 3 with CLIP-based matching; and (3) CLIP-Full: a fully retrieval-based pipeline where CLIP retrieves the correct procedure, matches the current step, and outputs the next step directly.

As shown in Table~\ref{table12}, all CLIP-based configurations underperform our method. While CLIP-P1 achieves relatively better results than other CLIP variants by leveraging LLM execution, it still shows a substantial gap compared to the baseline. These results demonstrate both the necessity of our proposed CoP and the inherent difficulty of this task. Even though the retrieval models can reasonably match the correct procedure, fine-grained alignment at the step level remains challenging.

\subsection{Human Evaluation}
To rigorously evaluate the effectiveness of our framework, we conduct a human evaluation comparing our method against two baselines: GPT-4o and a fine-tuned Qwen2.5-VL-7B. We randomly selected 20 instances from the benchmark dataset for these two VLMs separately and tasked three computer science students with pairwise comparisons. Annotators independently judged outputs based on accuracy (faithfulness to ground truth) and conciseness (brevity without redundancy), categorizing results as ``Better,'' ``Equivalent,'' or ``Worse'' relative to each baseline. 

As shown in Table \ref{table5}, our framework demonstrates statistically significant superiority across both metrics. Compared to the GPT-4o baseline, it achieves a 48\% ``Win'' rate with 20\% ``Equal'' rate. Against Qwen2.5-VL-7B baseline, it attains a 65\% ``Win'' rate, which is large greater than the ``Worse'' rate. Inter-annotator agreement (Fleiss $k=0.51$) confirms moderate consensus. These results validate that our CoP framework substantially improves both precision and clarity in next-step generation compared to existing approaches.

\begin{table}[t]
  \centering
    \renewcommand\arraystretch{1}
    \tabcolsep=0.32cm
    \scalebox{1}{
    \begin{tabular}{lccc}
    \toprule
     Model& Win & Equal & Loss   \\
      \midrule
      Qwen2.5-VL-7B & \textbf{65\%} & 2\% & 33\%  \\
      GPT-4o  & \textbf{48\%} & 20\% & 31\% \\
      \bottomrule
    \end{tabular}}
    \caption{The results of human evaluation on fine-tuned Qwen2.5-VL-7B and GPT-4o. \textbf{Bold} numbers indicate the better result for each VLM.}
  \label{table5}
\end{table}

\begin{figure}[t]
    \centering
    \includegraphics[width=1\linewidth]{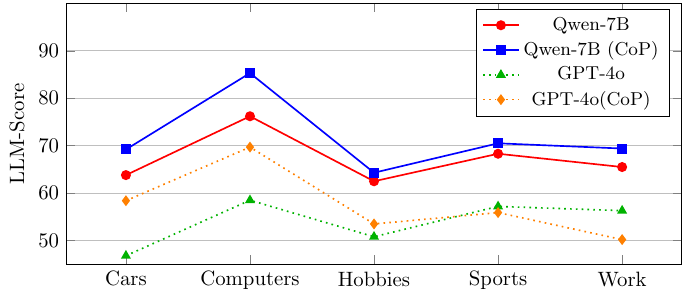}
    \caption{The LLM-score of fine-tuned Qwen2.5-VL-7B and GPT-4o on five domains separately.}
    \label{fig:5}
\end{figure}

\subsection{Performance Across Data Categories} \label{sec.6.5}
To analyze the robustness and generalization capabilities of CoP, we conduct experiments to compare its performance against baselines across different domains and step distribution. Figure \ref{fig:5} illustrates the results on five distinct domains (as described in Section \ref{sec.3.3}), where we employ LLM-Score as the evaluation metric due to its ability to jointly account for accuracy and conciseness. For the fine-tuned Qwen2.5-VL-7B, our framework consistently outperforms the baseline across all five domains. However, for GPT-4o, our approach underperforms the baseline in the ``Sports and Fitness'' and ``Work World'' domains. Considering that the fine-tuned Qwen2.5-VL-7B only marginally surpasses the baseline in these two domains, we think that ``Sports and Fitness'' and ``Work World'' procedures exhibit lower sensitivity to step order compared to the other three domains. Consequently, the second phase of our framework may introduce unintended noise or over-constraints, thereby negatively impacting final performance.

\begin{figure}[t]
    \centering
    \includegraphics[width=1\linewidth]{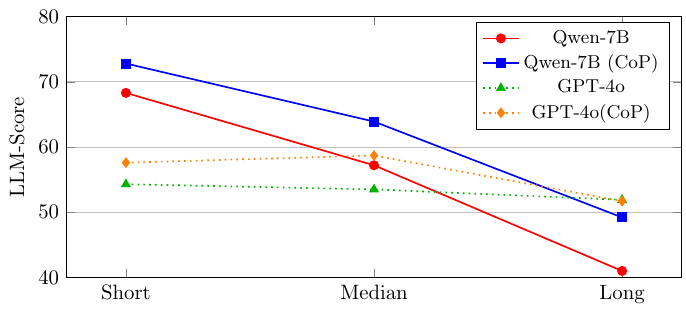}
    \caption{The average LLM-score of fine-tuned Qwen2.5-VL-7B and GPT-4o on different step lengths.}
    \label{fig:6}
\end{figure}

Figure \ref{fig:6} demonstrates the performance of our method versus baselines on procedures of varying step lengths. Our approach achieves superior results across all configurations. Empirically, procedures with more steps inherently demand higher comprehension complexity. These results suggest that powerful VLMs like GPT-4o can effectively handle difficult procedures in this task. In contrast, smaller-size VLMs such as Qwen2.5-VL-7B struggle to learn long procedural patterns. Nonetheless, our framework enhances their performance through systematic fine-tuning and inference following our proposed framework. Additionally, we compare the impact of different numbers of negative samples and sampling strategies in Appendix \ref{app.8}.

\subsection{Error Analysis} \label{sec.6.6}
\begin{figure}[t]
    \centering
    \includegraphics[width=1\linewidth]{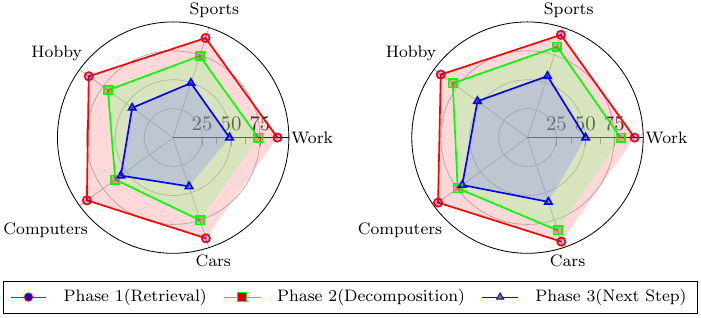}
    \caption{The results of three phases of our CoP on five domains on GPT-4o-mini (\textbf{Left}) and GPT-4o (\textbf{Right}).}
    \label{fig:8}
\end{figure}

To systematically evaluate the effectiveness of our method, we conduct an error analysis comparing model capabilities by implementing our three-phase framework on GPT-4o-mini and GPT-4o across five domains. Phase 1 performance is measured through procedure retrieval accuracy, Phase 2 through BLEU score alignment between decomposed and original procedures, and Phase 3 via LLM-Score of next-step prediction. The normalization scores are shown in Figure \ref{fig:8}.

As expected, GPT-4o consistently outperformed its smaller counterpart. Both models achieved over 90\% accuracy in Phase 1 procedure retrieval, demonstrating robust image-text alignment capabilities. However, significant performance gaps emerged in Phase 2 decomposition quality despite equivalent retrieved procedures, revealing critical differences in procedural understanding. This capability gap directly impacts Phase 3 outcomes in procedure-sensitive domains like ``Computers'', ``Cars'', and ``Hobby'', where GPT-4o's superior decomposition quality translates to 9.7-14.1\% higher LLM-Scores compared to GPT-4o-mini. Conversely, in knowledge-dependent domains, such as ``Sports'' and ``Work'', Phase 3 performance shows weaker correlation with decomposition quality, with GPT-4o achieving only marginal improvements over GPT-4o-mini in ``Work'' task. This explains the baseline performance gap observed in Figure \ref{fig:5} for these domains, suggesting that our method's effectiveness becomes constrained when procedural reasoning plays a secondary role to pre-existing world knowledge. The findings emphasize the dual dependence of visual procedural reasoning on both explicit procedure alignment and implicit procedural comprehension capabilities. More discussion will be shown in Appendix \ref{app.6}.

\section{Conclusion}
In this work, we investigate the underexplored capability of VLMs in the visual procedural reasoning task. We propose a novel multimodal benchmark designed for evaluating procedural reasoning. Through systematic analysis, we propose a CoP framework that synergizes cross-modal procedure retrieval, context-aware step decomposition, and the next step prediction. Experimental results show that our approach significantly outperforms baselines, revealing critical insights about cross-modal alignment and procedural reasoning. In the future, we will explore more efficient workflow adaptation for domain-specific procedures.

\section*{Limitations}
While our framework demonstrates significant improvements in visual procedural reasoning, we identify two key limitations. First, the multi-phase CoP framework costs around 2 times tokens compared to baselines, though our empirical results demonstrate that the substantial accuracy improvements justify this additional cost. The baseline costs average 2168 tokens per data, while our CoP method costs average 4526 tokens per sample. Second, the system remains susceptible to error propagation due to its dependence on off-the-shelf VLMs for procedure parsing during the retrieval stage. Notably, the performance is particularly sensitive to the accuracy of relevant procedure extraction, which critically impacts the final results.

\section*{Acknowledgments}
This work was supported in part by the Science and Technology Development Fund of Macau SAR (Grant Nos. FDCT/0007/2024/AKP, EF2024-00185-FST), the UM and UMDF (Grant Nos. MYRG-GRG2024-00165-FST-UMDF, MYRG-GRG2025-00236-FST), the Tencent AI Lab Rhino-Bird Research Program (Grant No. EF2023-00151-FST), the Stanley Ho Medical Development Foundation (Grant No. SHMDF-AI/2026/001), and the National Natural Science Foundation of China (Grant No. 62266013).

\bibliography{main}

\appendix

\section{Appendix}
\label{sec:appendix}

\subsection{Example of Different Domain Benchmark} \label{app.3}
To demonstrate the structure of our ProcedureVQA benchmark and the visual procedural reasoning task, Table \ref{table7} shows five examples covering the domains introduced in Section \ref{sec.3.2}. Each instance in our benchmark consists of a step-by-step procedural sequence, where the system must identify the correct next step given a query image. The default question is "What is the next step?", reflecting real-world use cases: when a user uploads an image showing a particular state, the QA system must retrieve the relevant procedure and predict the subsequent action.

\subsection{Sensitive Analysis of Fusion Probability} \label{app.9}
To justify our choice of the fusion probability parameter, we conducted a sensitivity analysis using GPT-4o, Gemini-2.5-Flash, and Claude-3.5-Sonnet on 1,000 sampled instances under three settings: Low-Fusion (25\%), Balanced (50\%), and High-Fusion (75\%). As shown in Table~\ref{tab:sensitivity}, BERTScore remains relatively stable across the three settings. These results validate the fusion probability as an effective difficulty control parameter, and our default setting of 50\% strikes a reasonable balance between sufficient concept integration and manageable complexity.

\begin{table}[t]
\centering
\begin{tabular}{llc}
\toprule
\textbf{Model} & \textbf{Setting} & \textbf{BERTScore}  \\
\midrule
\multirow{3}{*}{GPT-4o} & Low & 0.301  \\
& Balanced  & 0.305  \\
& High  & 0.312  \\
\midrule
\multirow{3}{*}{Gemini-2.5-Flash} & Low  & 0.453 \\
& Balanced & 0.463 \\
& High & 0.477  \\
\midrule
\multirow{3}{*}{Claude-3.5-Sonnet} & Low & 0.345  \\
& Balanced & 0.349  \\
& High  & 0.351  \\
\bottomrule
\end{tabular}
\caption{Sensitivity analysis of fusion probability across three LLMs on 1,000 sampled instances.}
\label{tab:sensitivity}
\end{table}

\subsection{Implementation} \label{app.2}
For closed-source VLMs and VLMs without fine-tuning, we directly use the models through their official APIs with temperature set to 0 to ensure experimental reproducibility. Regarding the fine-tuned Qwen2.5-VL-7B\footnote{https://huggingface.co/Qwen/Qwen2.5-VL-7B-Instruct} \cite{Qwen2VL} model described in Section \ref{sec.5}, we fine-tune (PEFT) it with LoRA \cite{hu2022lora}, a parameter-efficient fine-tuning method \cite{chen2025not} for LLMs, using the LLaMA-Factory\footnote{https://github.com/hiyouga/LLaMA-Factory} \cite{zheng2024llamafactory} framework on single H800 GPU. The training configuration employs a batch size of 4 with 4-step gradient accumulation, a learning rate of 1e-4, and a maximum sequence length of 8192 tokens. This setup ensures memory efficiency while maintaining stable optimization dynamics during the adaptation process. The two benchmarks constructed in this work will be released upon acceptance of this paper. 

\subsection{Detailed Sub-Task Formulations} \label{appendix:task_details}
We provide formal definitions for each diagnostic sub-task. 

\noindent\textbf{Shuffled Procedure Verification (SIV).} Given a procedure $\mathcal{P} = \{S_1, ..., S_L\}$, the model receives a permuted version $\mathcal{P}' = \{S_{\pi(1)}, ..., S_{\pi(L)}\}$ where $\pi$ is a random permutation. The task is binary classification: predict whether $\mathcal{P}' = \mathcal{P}$. This tests sensitivity to procedural order without requiring visual grounding.

\noindent\textbf{Current Step Identification (CSI).} Given a visual query $v_t$ and the complete procedure $\mathcal{P} = \{S_1, ..., S_L\}$, the model identifies which step $S_i \in \mathcal{P}$ corresponds to the current visual state (multi-class classification over $L$ steps). This evaluates temporal grounding between vision and text.

\noindent\textbf{Next-Step Prediction (NSP).} Given visual query $v_t$ showing the state after completing $S_t$ and the procedure $\mathcal{P}$, the model predicts the next step $S_{t+1}$. Unlike CSI, which requires recognition, NSP demands forward temporal reasoning about future actions.

\noindent\textbf{Disordered Procedure Adaptation (DPA).} The model receives visual query $v_t$ and shuffled procedure $\mathcal{P}' = \{S_{\pi(1)}, ..., S_{\pi(L)}\}$, then must predict the correct next step $S_{t+1}$ from the original sequence $\mathcal{P}$. This simulates real-world scenarios where provided procedures may be incomplete or misordered.

\noindent\textbf{Cross-Procedure Matching (CPM).} Given a visual query $v_t$ depicting procedure $\mathcal{P}_i$ and a candidate procedure description $\mathcal{P}_j$, the model determines if $\mathcal{P}_i = \mathcal{P}_j$ (binary classification). This tests cross-modal semantic alignment without requiring fine-grained temporal reasoning.

Each task uses identical image-procedures pairs from our benchmark but varies the input format and prediction target to isolate specific reasoning capabilities. Note that these decomposed sub-tasks simplify the full VP-QA problem by removing one or more of the three core challenges: cross-procedure retrieval, visual state grounding, and next-step reasoning.

\subsection{Evaluation Metrics} \label{app.1}
The LLM-Score employs Claude-3.7 \cite{claude-3.7}, Gemini-2.5 \cite{DBLP:journals/corr/abs-2312-11805}, and GPT-4.1 \cite{gpt-4.1} as expert judges following \citet{chia2024m}. These LLMs independently assess the semantic equivalence between model outputs and reference next-step actions through a scoring prompt. Final LLM-Scores represent normalized averages across all three judges, mitigating individual model biases. The instructions of all LLM judge experts are shown in Table \ref{table6}.

\begin{figure}[t]
    \centering
    \includegraphics[width=1\linewidth]{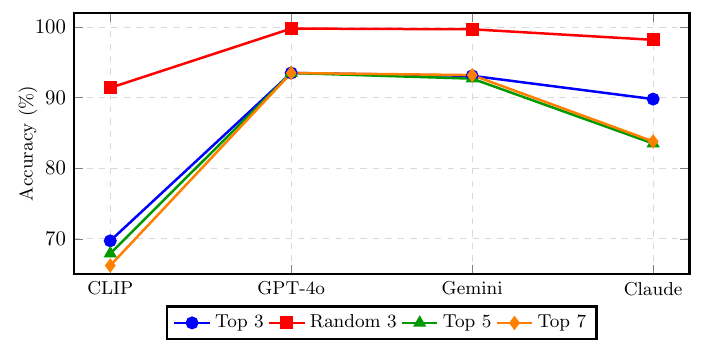}
    \caption{The results with different numbers of negative candidate procedures and sampling strategies.}
    \label{fig:9}
\end{figure}

\subsection{The Impact of Negative Manuals} \label{app.8}
We analyze Phase 1 matching accuracy under different scenarios in Figure \ref{fig:9}, varying the number of negative candidate procedures and sampling strategies across retrieval models and vLLMs (CLIP, GPT-4o, Gemini-2.5-Flash, and Claude-3.5-Sonnet). Top-k (k=3,5,7) refers to our construction in Section \ref{sec.3.2}, where candidates include one correct procedure and N-1 visually similar negative procedures; Random-3 uses one correct procedure with two randomly sampled negatives. The results reveal two key findings: First, Random-3 achieves substantially higher accuracy than all Top-k settings, demonstrating the effectiveness of our hard negative sampling strategy. Second, increasing the number of negatives consistently degrades performance on CLIP and Claude-3.5-Sonnet, with slight drops even on GPT-4o and Gemini-2.5-Flash, highlighting the necessity and rationale of using large language models rather than pure matching models for Phase 1.

\subsection{Instruction Template of VLMs} \label{app.5}
This section presents the complete collection of instructions used throughout our experimental analysis. As shown in Table \ref{table8}, we provide the standardized prompt for our VLM baselines, where VLMs directly predict the next step. Table \ref{table9} outlines the operationalization of our five evaluation subtasks designed to assess VLM capabilities. Finally, Table \ref{table10} details the phase-specific instructions corresponding to different phases of our proposed CoP framework.

\subsection{Discussion of Error Propagation} \label{app.6}
By comparing the two plots in Figure \ref{fig:8}, we observe that for VLMs of different sizes, larger models consistently achieve better performance in phase 2, which subsequently leads to improved performance in phase 3. However, for the same VLM across different domains, the performance of phase 2 and phase 3 does not exhibit a positive correlation, indicating that the model possesses certain robustness against error propagation throughout the pipeline. This robustness explains why fine-tuned models demonstrate substantial overall improvement, as capabilities are enhanced across all three phases. Nevertheless, examining the performance variations of the three phases across five domains reveals that our proposed method can better leverage the model's inherent capabilities, ultimately achieving better performance on the final task compared to the baseline.

\subsection{Synthesize Fine-tuned Data} \label{app:7}
We leverage Qwen3-VL-235B-A22B-Thinking~\cite{qwen3technicalreport} for synthesizing procedure-aligned reasoning chains, following the data construction protocol outlined in Section~\ref{sec.3}. The synthesis process involves presenting the model with candidate procedures alongside query images, then explicitly specifying: (1) the appropriate procedure selection, (2) whether finer-grained decomposition is necessary, (3) the step corresponding to the query image, and (4) the subsequent action to execute. Based on these structured inputs, the model generates complete thinking trajectories. A subsequent LLM-based verification stage examines both step-wise answer correctness and adherence to the prescribed information sequence. Failed samples undergo up to three regeneration attempts before human annotators take over to craft the reasoning process manually, ensuring consistent quality across our fine-tuning and benchmark datasets.

\begin{table*}[t]
\centering
\begin{tabular}{lll}
\toprule
\multicolumn{1}{c}{Cars \& Other Vehcles} & \multicolumn{1}{c}{Computers and Electronics} & \multicolumn{1}{c}{Hobbies and Crafts} \\
\midrule
\textbf{Procedure 1:} & \textbf{Procedure 1:} & \textbf{Procedure 1:}\\
Step 1: Pour coolant into the ... & Step 1: Understand shortcut ... & Step 1: Purchase frisket film ... \\
... (Other Steps) & ... (Other Steps) & ... (Other Steps) \\
Step 4: Top off your coolant ... & Step 10: Back up your files ... & Step 6: Remove frisket from ... \\
\textbf{(Other extracted procedures)} & \textbf{(Other extracted procedures)} & \textbf{(Other extracted procedures)} \\
\\
\textbf{User Query Image:} & \textbf{User Query Image:} &  \textbf{User Query Image:} \\
\includegraphics[width=0.3\textwidth]{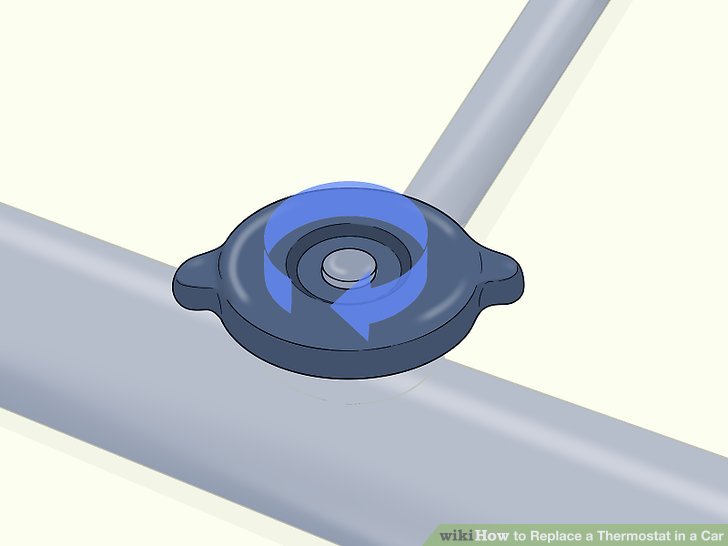} & \includegraphics[width=0.3\textwidth]{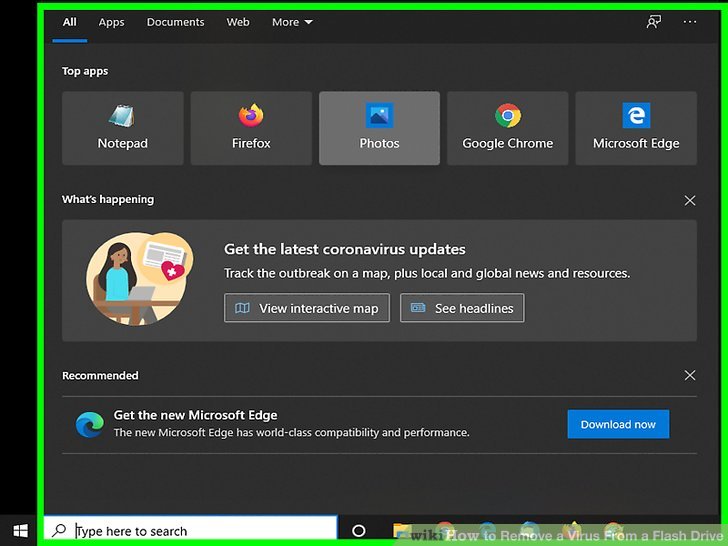} & 
\includegraphics[width=0.3\textwidth]{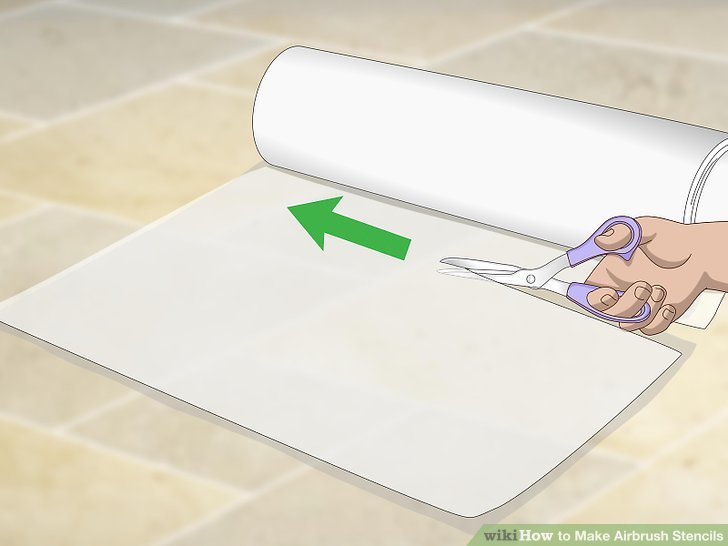}  \\
\textbf{\color{red}{Golden Next Step:}} & \textbf{\color{red}{Golden Next Step:}}  & \textbf{\color{red}{Golden Next Step:}} \\
Start your vehicle’s engine ... & Type command prompt. A ... & Place unpeeled frisket over ... \\
\midrule
\midrule
\multicolumn{1}{c}{Sports and Fitness} & \multicolumn{1}{c}{Work World} & \\
\midrule
\textbf{Procedure 1:} & \textbf{Procedure 1:} & \\
Step 1: Pick skis that are ... & Step 1: Let them lead in the ... &  \\
... (Other Steps) & ... (Other Steps) & \\
Step 6: Protect yourself. ... & Step 3: Listen for sincerity. ... &  \\
\textbf{(Other extracted procedures)} & \textbf{(Other extracted procedures)} &  \\
\\
\textbf{User Query Image:} & \textbf{User Query Image:} & \\
\includegraphics[width=0.3\textwidth]{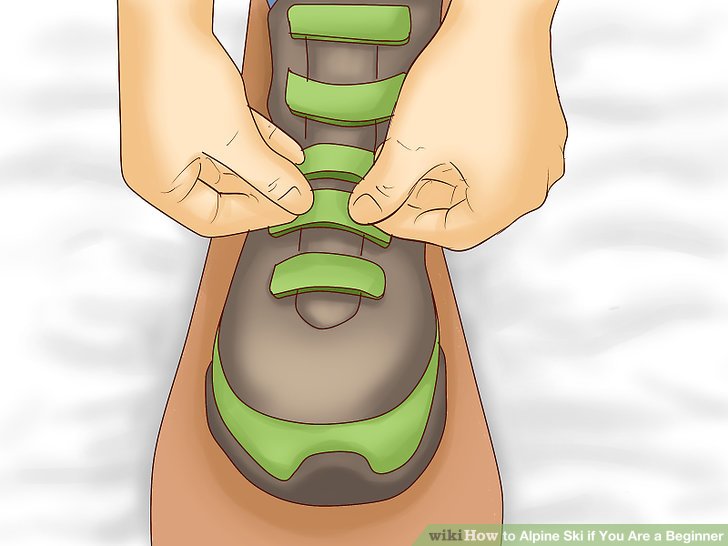} & \includegraphics[width=0.3\textwidth]{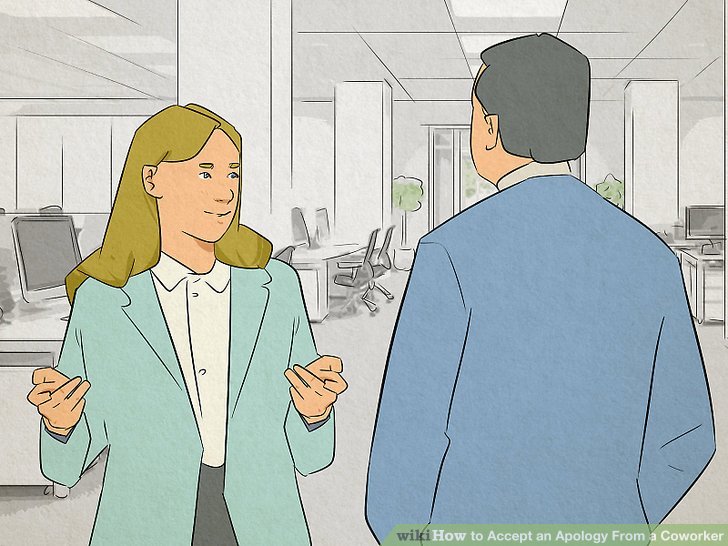} &   \\
\textbf{\color{red}{Golden Next Step:}} & \textbf{\color{red}{Golden Next Step:}}  & \\
Grab a set of poles. Ski poles ... & Don’t brush off their guilt. ... &  \\
\bottomrule
\end{tabular}
\caption{The examples of five domain data in our benchmark.}
\label{table7}
\end{table*}

\begin{table*}[h]
  \centering
    \renewcommand\arraystretch{1}
    \scalebox{1}{
    \begin{tabular}{l}
    \toprule
    \multicolumn{1}{c}{Instruction of LLM-Score} \\
    \midrule
    \#\# Correct Next Step: \\
    \{LABEL\} \\
    \\
    \#\# Predict Next Step: \\
    \{PREDICT\} \\
    \\ 
    \#\#Instructions: \\
    - The ``Correcct Next Step'' is the ground truth of the next step. \\
    - The ``Predict Next Step'' is the model output of the next step. \\
    - Score the predicted next step (0-10) based on how concisely and accurately it covers all essential \\ operations from the correct next step, without omissions or extra information. \\
    - The score should be between 0 to 10. \\
    - Please output the score directly without any introduction or explanation. \\
      \bottomrule
    \end{tabular}}
    \caption{The instruction for the LLM judge expert to score the experimental outputs.}
  \label{table6}
\end{table*}

\begin{table*}[t]
  \centering
    \renewcommand\arraystretch{1}
    \scalebox{1}{
    \begin{tabular}{l}
    \toprule
    \multicolumn{1}{c}{Instruction of VLM baselines} \\
    \midrule
    \#\# Instruction Manuals: \\
    \{STEPS\} \\
    \\
    \#\#Instructions: \\
    - Several instructions are provided, each containing steps in sequential order (with no shuffling). \\
    - Some steps may combine multiple actions into a single step. \\
    - Based on the given image, identify the corresponding instruction manual and determine the next \\ step. \\
    - Please think step-by-step and make sure the last line of the output should be the following format: \\
     step\_X: [content of the next step] \\
      \bottomrule
    \end{tabular}}
    \caption{The instruction for VLM baselines.}
  \label{table8}
\end{table*}

\begin{table*}[t]
  \centering
    \renewcommand\arraystretch{1}
    \scalebox{1}{
    \begin{tabular}{l}
    \toprule
    \multicolumn{1}{c}{Instruction of Predict Current Step} \\
    \midrule
   <Image></Image> \\
    \\
    \#\# Instructions: \\
    \{STEPS\} \\
    \\
    - Please identify which step the image corresponds to, based on the given instruction. \\
    - You should only output text in the form of 'step\_X', without any other explanations or descriptions. \\ 
    \midrule
    \multicolumn{1}{c}{Instruction of Predict Next Step} \\
    \midrule
    <Image></Image> \\
    \\
    \#\# Instructions: \\
    \{STEPS\} \\
    \\
    - Please identify the next step after the step shown in the given image, based on the given instruction. \\
    - You should only output text in the form of 'step\_X', without any other explanations or descriptions. \\ 
    \midrule
    \multicolumn{1}{c}{Instruction of Randomly Shuffled Detection} \\
    \midrule
    \#\# Instructions: \\
    \{STEPS\} \\
    \\
    - Please determine whether the operational procedure in the instructions has been shuffled. \\
    - You should only output 'True' or 'False', without any other introduction or explanation.  \\
    \midrule
    \multicolumn{1}{c}{Instruction of Predict Next Step in Shuffled Instructions} \\
    \midrule
    <Image></Image> \\
    \\
    \#\# Instructions: \\
    \{STEPS\} \\
    \\
    - Given the shuffled instructions, please identify the next step after the step shown in the given image. \\
    - You should only output text in the form of ``step\_X'', without any other introduction or explanation.  \\
    \midrule
    \multicolumn{1}{c}{Instruction of Cross-Procedure Matching} \\
    \midrule
    <Image></Image> \\
    \\
    \#\# Instructions: \\
    \{STEPS\} \\
    \\
    - Please determine whether the content of the image corresponds to any step described in the given \\ instruction. \\
    - You should only output 'True' if the image corresponds to the instruction, or 'False', without any \\ other introduction or explanation.  \\
      \bottomrule
    \end{tabular}}
    \caption{The instructions for the five evaluation sub-tasks.}
  \label{table9}
\end{table*}

\begin{table*}[t]
  \centering
    \renewcommand\arraystretch{1}
    \scalebox{1}{
    \begin{tabular}{l}
    \toprule
    \multicolumn{1}{c}{Instruction of Context-Aware Procedure Retrieval} \\
    \midrule
    <Image></Image> \\
    \\   
    \#\# Task: Identify which instruction manual the image belongs to. \\
    \\
    \#\#Instructions: \\
    \{STEPS\} \\
    \\
    - Several instructions are provided (start with \#\#\#), each containing steps in sequential order \\ (with no shuffling). \\
    - Based on the given image, identify the corresponding instruction among the provided candidate \\ instructions. \\
    - Ensure that your response includes only the instruction ID. \\
    - Output your result in the following format: [Instruction id]\\
    - Do not include any explanations, reasoning, or additional information in your response.\\
    \midrule
    \multicolumn{1}{c}{Instruction of Step-Wise Procedural Decomposition}\\
    \midrule
    \#\# Task: Identify and split the combined steps in the instruction manual. \\
    \\
    \#\# Instruction: \\
    \{STEPS\} \\
    \\
    - The instruction contains steps in strict sequential order (no shuffling or reordering). \\
    - Some steps may combine two distinct actions into a single step. \\
- Split if the actions are clearly separate (unrelated tools, different targets, or independent \\ operations). \\
- Do not split if the actions are part of a continuous process (same tool/object,\\ sequential dependencies, or a single logical operation). \\
- Output the modified instruction with only the necessary splits applied, keeping all other steps and \\the step description unchanged. \\
- Output your result in the following format: \\
    step\_1: [content of step\_1] \\
    step\_2: [content of step\_2] \\
    ... \\
    step\_X: [content of step\_X] \\ 
- Do not include any explanations, reasoning, or additional information in your response. \\
        \midrule
         \multicolumn{1}{c}{Instruction of Next Step Prediction}\\
         \midrule
        <Image></Image> \\
         \\
         \#\# Task: Identify which step in the instruction the input image belongs to. \\
        \\
        \#\# Instruction:\\
        \{STEPS\} \\
        \\
    
        - The instruction contains steps in sequential order (with no shuffling). \\
        - Based on the given image, identify which step this image belongs to. \\
        - Ensure that your response includes only the step of the image. \\
        - Output your result in the following format: \\
            step\_X: [content of the current step] \\
        - Do not include any explanations, reasoning, or additional information in your response.\\
      \bottomrule
    \end{tabular}}
    \caption{The instructions for each phase in our CoP framework.}
  \label{table10}
\end{table*}

\end{document}